\documentclass[sigconf]{aamas}

\usepackage{balance} % for balancing columns on the final page
\usepackage{algorithm}
\usepackage{algorithmic}
\usepackage{multirow}
\usepackage{comment}
\usepackage{cleveref}

\usepackage{amsmath}

\DeclareMathOperator*{\argmin}{arg\,min}

\newcommand\ben[1]{\textcolor{blue}{Ben says: #1}}
\newcommand\carter[1]{\textcolor{purple}{Carter says: #1}}

\setcopyright{none}
\acmConference[SCaLA-24]{Appears at the 1st Workshop on Social Choice and Learning Algorithms (SCaLA 2024) held at the 23rd International Conference on Autonomous Agents and Multiagent Systems.}
{May 6-7, 2024}{Auckland, New Zealand}{Armstrong, Fairstein, Mattei, Terzopoulou}
\copyrightyear{2024}
\acmYear{2024}
\acmDOI{}
\acmPrice{}
\acmISBN{}
\acmSubmissionID{8}

\title[Liquid Ensemble Selection for Continual Learning]{Liquid Ensemble Selection for Continual Learning}

\author{Carter Blair}
\affiliation{
  \institution{University of Waterloo}
  \country{Canada}
  }
\email{carter.blair@uwaterloo.ca}

\author{Ben Armstrong}
\affiliation{
  \institution{University of Waterloo}
  \country{Canada}
  }
\email{ben.armstrong@uwaterloo.ca}

\author{Kate Larson}
\affiliation{
  \institution{University of Waterloo}
  \country{Canada}
  }
\email{kate.larson@uwaterloo.ca}

\begin{abstract}

Continual learning aims to enable machine learning models to continually learn from a shifting data distribution 
without forgetting what has already been learned. 
Such shifting distributions can be broken into disjoint subsets of related examples; by training each member of an ensemble on a different subset it is possible for the ensemble as a whole to achieve much higher accuracy with less forgetting than a naive model. We address the problem of selecting which models within an ensemble should learn on any given data, and which should predict.
By drawing on work from delegative voting we develop an algorithm for using delegation to dynamically select which models in an ensemble are active. We explore a variety of delegation methods and performance metrics, ultimately finding that delegation is able to provide a significant performance boost over naive learning in the face of distribution shifts.

\end{abstract}


\newcommand{\BibTeX}{\rm B\kern-.05em{\sc i\kern-.025em b}\kern-.08em\TeX}

\begin{document}

%%% The following commands remove the headers in your paper. For final 
%%% papers, these will be inserted during the pagination process.

\pagestyle{fancy}
\fancyhead{}

%%% The next command prints the information defined in the preamble.

\maketitle 

%%%%%%%%%%%%%%%%%%%%%%%%%%%%%%%%%%%%%%%%%%%%%%%%%%%%%%%%%%%%%%%%%%%%%%%%

\section{Introduction}

Machine learning models often operate under the assumption that the data they are trained on and the data on which they will make predictions are independently and identically distributed (i.i.d.). However, this assumption does not reflect the reality of the world we live in. In practical settings, data distributions are constantly shifting. Deployed machine learning models that operate as though the world is static will eventually become highly inaccurate or will begin to enforce outdated knowledge about the world -- often resulting in harmful bias. Instead, models must adapt to the world as it changes.

\textit{Continual learning} is a machine learning framework in which models learn incrementally on training data that arrives in an ordered stream and from a shifting distribution. Naively learning on such data allows models to remain performant on recent data but leads to a problem known as \textit{catastrophic forgetting} wherein previously learned information is forgotten \cite{french1999catastrophic, van2022three}. Similarly, test data is often modelled as coming from a similar non-stationary stream where changes in distribution typically harm model performance. Work in continual learning develops more reactive learning algorithms in order to address problems caused by distribution shifts \cite{van2022three}.

Ensembles have long been used in many machine learning settings. In standard learning situations, ensembles typically outperform single classifiers and demonstrate a strong ability to leverage diverse models for a common purpose. Conceptually, ensembles are well-suited to data shifts:
By using different classifiers to learn on different underlying distributions the ensemble can dynamically select classifiers to rely upon when making predictions and avoid forcing a single model to forget previous knowledge in order to stay up-to-date. Existing work has already found that ensembles can be applied to a continual learning setting \cite{doan_continual_2023, rypesc_divide_2024, wang_coscl_2022, wen_batchensemble_2020, aljundi_expert_2017, liu_more_2020}.

We argue that the recent use of delegative voting for ensemble pruning lays a foundation for a new continual learning algorithm.
This delegative voting framework, known as \textit{liquid democracy}, allows ensemble members to participate directly in training and inference, or to transitively delegate their action to another classifier. 

While many continual learning algorithms rely upon explicit knowledge about changes in the distribution \cite{rypesc_divide_2024, aljundi_expert_2017}, or remember useful training examples so they may be re-learned  \cite{pan_continual_2020, chaudhry2018efficient, rebuffi_icarl_2017}, our approach relies only upon knowledge about the recent learning performance of each classifier. Using this information, the algorithm makes delegation decisions which further two goals: (1) selecting the classifiers most well suited to learning/predicting under the current distribution and (2) weighting the chosen classifiers appropriately in order to optimize their prediction accuracy.

Our algorithm is highly generic; we explore a variety of delegation methods drawn from existing research on liquid democracy and several measures of classifier performance. Rather than relying solely on accuracy, we are able to consider any single-valued measure such as precision, recall or f1-score. In experiments across multiple domains of continual learning, we demonstrate the utility of adapting ideas from liquid democracy into the continual learning setting. In particular, our contributions are as follows:

\begin{itemize}
    \item A novel framework for continual learning and dynamic ensemble selection based on principles of liquid democracy.%, addressing the limitations of traditional learning methods in adapting to non-stationary environments.
    \item New definitions of classifier performance in terms of learning rate and recent performance, tailored to the specific demands of continual learning.
    \item Empirical validation of our approach via experiments on classic continual learning benchmarks, demonstrating a strong ability to enhance ensemble performance.
\end{itemize}

Through these contributions, our work establishes a new direction for research in continual learning and dynamic ensemble selection. It highlights the potential of integrating concepts from delegative voting into machine learning to create more adaptable and diverse ensembles.

\section{Related Work}

\subsection{Ensembles in Social Choice}

Ensembles rely on aggregation methods to combine outputs from many classifiers into a single prediction. As such, a great deal of work spanning multiple decades has considered the application of social choice results to ensemble learning. This has been illustrated through theoretical connections between ensemble learning and well-known axiomatic results in social choice \cite{pennock2000normative}, as well as by experimental evaluation of many common voting rules for classifier prediction aggregation \cite{cornelio2021voting,leon2017evaluating}.
% While not directly studying machine learning, \citeauthor{christoff2017binary} have a setting highly similar to binary classification -- epistemic binary aggregation -- and shown that delegation in their setting is often not rational for individual voters.

Only one line of work has considered the application of liquid democracy to ensembles. In a setting with exactly one round of delegation, liquid democracy has been shown to not provide any significant benefit over traditional ensembles \cite{armstrong2021limited}. However, delegation can also be viewed as a form of ensemble pruning and reweighting. Using iterative delegations in an incremental learning setting, \citeauthor{armstrong2024liquid} showed that liquid democracy can typically achieve higher accuracy than a full ensemble while also dramatically reducing the cost of training the ensemble \cite{armstrong2024liquid}. On some datasets, their method attained higher accuracy than the well-known ensemble algorithm Adaboost \cite{freund1995decision}. However, this result was highly data-dependant.
While the incremental learning setting of \citeauthor{armstrong2024liquid} (\citeyear{armstrong2024liquid}) can be considered a continual learning setting, there appears to be no work 
applying social choice to continual learning in a setting with distributional shift.

\subsection{Ensembles in Continual Learning}

Ensembles are a useful method to improve performance in continual learning~\cite{doan_continual_2023, rypesc_divide_2024, wang_coscl_2022, wen_batchensemble_2020, aljundi_expert_2017, liu_more_2020}.
On one extreme end of the spectrum, some ensemble methods initialize a new member for each new context they encounter \cite{aljundi_expert_2017}. However, these methods have high resource usage and require context labels in training, limiting their applicability. Other methods, such as SEED \cite{rypesc_divide_2024} and CoSCL \cite{wang_coscl_2022}, use a fixed number of models in the ensemble to avoid a linear increase in parameters with the number of contexts. These methods also require the context label to be known during training.

In contrast to other methods, we focus on keeping the architecture simple and general. We do not use a generative model, replay buffer or leverage context labels in training, and we use a fixed number of models in our ensemble to limit parameter growth. Not relying on context labels in training allows our method to work in a more diverse range of settings, including both class-incremental and domain-incremental learning settings.

%%%%%%%%%%%%%%%%%%%%%%%%%%%%%%%%%%%%%%%%%%%%%%%%%%%%%%%%%%%%%%%%%%%%%%%%

\section{Model}

Our model bridges two fields of study -- machine learning, and social choice. We consider an epistemic social choice setting where voters are instantiated as an ensemble of machine learning classifiers. Voters (classifiers) may vote for one of several alternatives in pursuit of a singular ``correct'' alternative and this corresponds to the classification action of a classifier. 

We focus on a setting where data has not only features and labels, but also a context which is associated with some data distribution. Within this setting, we explore algorithms for dynamically re-weighting voters/classifiers using delegative techniques. That is, classifier weights are controlled by ``delegation'' between classifiers, and only non-delegating classifiers take part in inference or learning. Classifiers may participate actively (they learn/predict on examples directly) or may delegate to another classifier, not thereby increasing that classifiers prediction weight and not learning or predicting.

In more detail, we consider an ensemble of classifiers $V = \{v_1, v_2,$ $..., v_n\}$, which we equivalently refer to as voters, and a non-stationary data stream $\mathcal{D}$ containing $m$ classes. Data arrives in batches of $B$ samples over $T$ steps, with each batch associated with some (possibly repeating) context $c_t \in \mathcal{C}$; $\mathcal{D} = \{(\mathcal{X}_1, \mathcal{Y}_1, c_1), ..., (\mathcal{X_T}, \mathcal{Y_T}, c_T)\}$. Each example in a dataset can be decomposed into three pieces: features $\mathcal{X}$, a class label $\mathcal{Y}$, and a context label $\mathcal{C}$. Together, the class label and context label make up a \textit{global label} $\mathcal{G}$. Depending upon the setting, this stream may include training data, testing data, or training data \textit{and} testing data. \autoref{fig:learning_target_explanation} gives an example of the difference between a class label and a context label.

A delegation function, $d: V \times T \rightarrow V$, determines which classifiers are learning or doing inference on any given batch. A delegation function value $d(v_i, t) = v_j$ indicates that, during batch $t$, $v_i$ delegates to $v_j$. In most cases, the batch is clear from context, and we write $d(v_i) = v_j$ to indicate a delegation in the current batch. By default, classifiers delegate to themselves. Classifiers that delegate to anyone other than themselves are referred to as ``delegators'' and are \textit{inactive} while a classifier that delegates to itself is \textit{active} and may be referred to as a ``guru.'' The set of all gurus is found through repeated application of the delegation function until a self-delegation is reached: $G_t(V) = \{v \in V| d^*(v, t) = v\}$.
Only classifiers that are gurus participate in learning or inference, depending on the setting.

The weight of each classifier is also controlled by delegation. Specifically, a weight vector $W_t$ contains the weight associated with each classifier on batch $t$. If no classifiers delegate (that is, if they all delegate to themselves), then each classifier has a weight of 1. Generally, each classifier has a weight equal to the number of delegations they receive, including their own:

\[
W_{t}[i] = 
\begin{cases}
0 & \text{if } v_i \notin G_t(V) \\
|\{v_j \in V| d^*(v_j, t) = v_i\}| & \text{otherwise }
\end{cases}
\]

An estimate of the performance of each classifier is continually refined. Performance can be measured as prediction accuracy or any other single-valued metric that reflects the performance of a classifier on a batch of data. A = $\lbrack a_{t, i}\rbrack$ stores the performance of each classifier $v_i$ on each batch $t$.
We are often interested in the trend of a classifier's performance as it learns. That is, over recent batches, has it become more or less accurate? We use this to approximate two measures: (1) whether or not the classifier is still improving its performance/benefits from continued training, and (2) as a proxy for whether a context shift has happened. If a classifier's performance drops rapidly, this is an indication that the underlying data has shifted distributions.
To evaluate this performance trend, we calculate the slope of the linear regression line of the performance values of each classifier over some parameter $w$ of recent batches. We denote the performance slope of $v_i$ as $q_{i, w}$ and all such slopes as $Q$. The set of all classifiers with a higher slope than $v_i$ is $N^+(v_i) = \{v_j \in V | q_{j,w} > q_{i,w}\}$. Intuitively, this corresponds to the classifiers that are ``better'' than $v_i$; those showing more recent improvement in accuracy.
Typically, we refer to the current weight and estimated performance of classifier $v_i$ as $w_i$ and $q_i$, respectively, only including a time subscript where necessary.

To generate a prediction, each guru in the ensemble generates class probabilities for each possible class. These are given a weight equal to the guru's weight and summed to get a weighted probability for each class. The class with the highest weighted probability is the prediction of the ensemble. This procedure is based on that of the SEED method from \citeauthor{rypesc_divide_2024} \cite{rypesc_divide_2024}.

% \todo{update; no inference}
% Throughout this paper we focus on two separate problems: (1) \textit{inference} given pretrained classifiers, and (2) \textit{learning} in a continual learning framework.

% \subsection{Inference}

% % \begin{itemize}
% %     \item A set of pretrained classifiers
% %     \item A non-stationary (non-i.i.d.) test stream
% %     \item only manipulate testing and only care about test performance
% % \end{itemize}

% \ben{Are we reporting results on this setting?}

% In this setting, we begin with \textit{fully trained} classifiers. Each classifier is trained on some subset of class labels (possibly all labels). $\mathcal{D}$ consists of batches of test data and the goal is to maximize ensemble prediction accuracy. $\mathcal{D}$ remains non-stationary; the distribution of class labels may be different in each context. Performance is improved by using delegation to empower voters with higher performance on a particular context. We explore a variety of delegation methods that work toward this goal.

\subsection{Continual Learning}

\begin{figure*}
    \centering
    \includegraphics[width=0.8\textwidth]{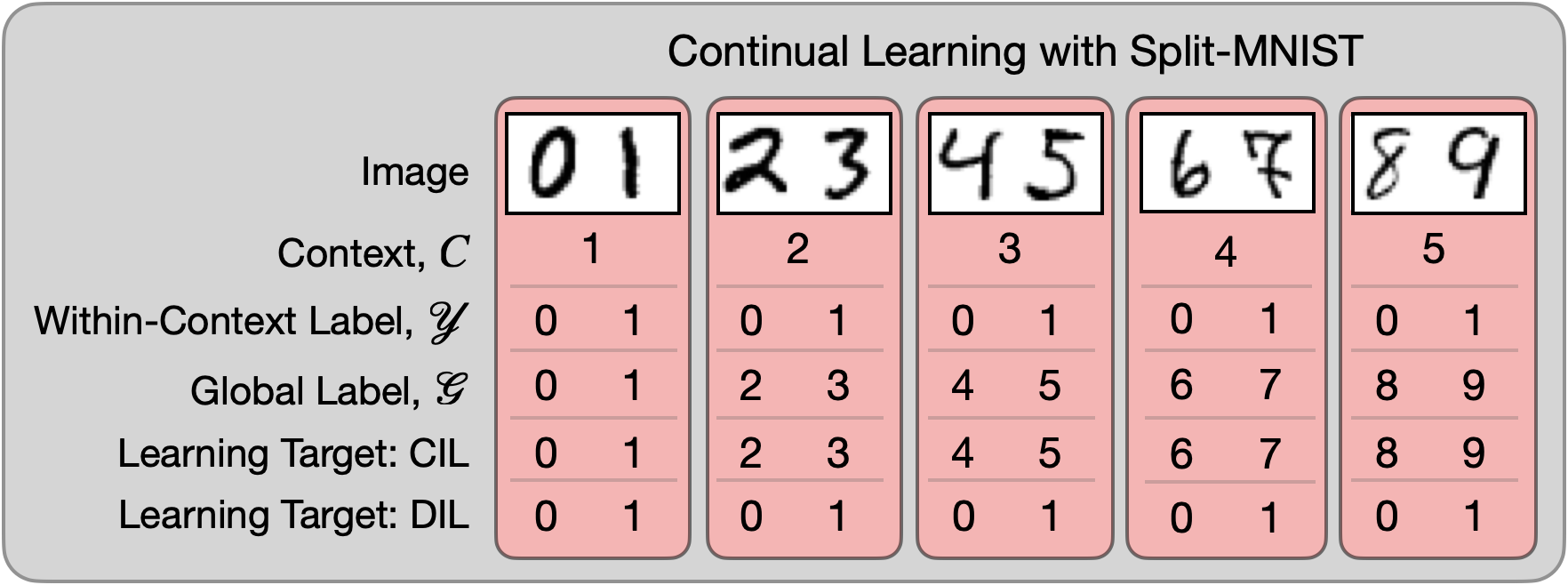}
    \caption{Examples of label types and learning targets within continual learning. Classes are grouped into disjoint contexts which are concatenated to form a data stream. Class Incremental learning learns the global label -- a unique combination of context label and within-context label. Domain Incremental learning learns only the label within the context. (Note: This figure is inspired by a similar figure from \citeauthor{van2022three} \cite{van2022three}.)}
    \label{fig:learning_target_explanation}
\end{figure*}

In continual learning, a learner attempts to learn from a non-stationary stream of data arriving in batches, each associated with some context. A context may appear in multiple contiguous batches, but, importantly, after a context stops arriving, it does not return during the training phase. This can lead to what is commonly referred to as ``catastrophic forgetting,'' which is characterized by the learner forgetting how to classify samples from contexts learned early in training after further learning in other contexts has occurred. 

We explore two of the standard scenarios within continual learning, differentiated based upon the information available to the learner and the label space~\cite{wang2023comprehensive,van2022three}:

\begin{enumerate}
    % \item In \textit{task-incremental learning}, contexts have disjoint class labels. Classifiers know the current context and which labels exist within each context during training and at test time. Formally, a function $f$ is learned where $f:\mathcal{X} \times \mathcal{C} \rightarrow \mathcal{Y}$. \ben{Probably remove this?}
    \item In \textit{class-incremental learning}, contexts have disjoint class labels. Classifiers only know the current context during training, not testing. The classifier must learn to predict both the label and context given only the input. Formally, the goal is to learn a mapping $f:\mathcal{X} \rightarrow \mathcal{Y}\times \mathcal{C}$.
    \item In \textit{domain-incremental learning}, each context has the same class labels, but they may appear at different frequencies in different contexts. The goal is to learn a mapping $f:\mathcal{X}\rightarrow\mathcal{Y}$.
\end{enumerate}

\autoref{fig:learning_target_explanation} gives an example of the learning target of both learning methods. \autoref{alg:dynamic_lde} outlines the general behaviour of our continual learning training algorithms: Voters all start as gurus, then, for each batch of data, the ensemble receives the batch, has all gurus learn the batch, and updates delegations based on training performance.
Note that in all settings, we assume the training data is ordered or non-stationary, but there are no assumptions about the test set. We can construct the test set as useful for evaluation. For example, it is common to report test accuracy over all contexts and within context. We may assume that the test set is non-stationary in some cases.

\begin{algorithm}[tb]
\caption{Continually Learning Liquid Democracy Ensemble}
\label{alg:dynamic_lde}
\begin{algorithmic}
\STATE {\bfseries Input:} Batch size $B$, number of classifiers $n$, delegation mechanism $DelMech$
\STATE Initialize all classifiers $\{v_i\}_{i=1}^{n}$ and all delegations to $d(v_i) = v_i$

    \FOR{t = 1, ..., T}
    \STATE Receive batch $\{(x_{j}, y_{j})\}_{j=1}^{B}$ from data stream
        \FOR{$v_i \in V$}
            \STATE $v_i$ makes predictions $\{\hat{y}_{i,j}\}_{j=1}^{B}$ on current batch
            \IF{$ d(v_i) = v_i$ (classifier is guru)}
                \STATE $v_i$ learns on batch
            \ENDIF
        \ENDFOR
        \STATE Update delegations: $DelMech(V)$
        
    \ENDFOR
\end{algorithmic}
\end{algorithm}

%%%%%%%%%%%%%%%%%%%%%%%%%%%%%%%%%%%%%%%%%%%%%%%%%%%%%%%%%%%%%%%%%%%%%%%%

\section{Delegation Mechanisms}

We develop a delegation mechanism suitable for continual learning that aims to be adaptive to context shifts while incorporating existing delegation methods as a parameter. Our mechanism, $k$-Best-Accuracy-Trend ($k$-BAT), does not depend on knowledge about the current context or store previous examples. Rather, the mechanism uses a weak signal about the recent performance of each classifier in the ensemble to determine whether/where each classifier should delegate.

\subsection{$k$-Best-Accuracy-Trend Delegations} \label{k-BAT}

Our algorithm is described in \autoref{alg:dynamic_bat} and here in words.
In continual learning, data is organized into contexts which are further subdivided into batches. $k$-BAT works over batches, using two parameters to record performance and make decisions: A \texttt{metric} parameter evaluates the performance of classifiers and a \texttt{delegation probability function} determines where classifiers will delegate.

During learning, $k$-BAT is called once per batch after the active voters in the ensemble have learned on the batch. 
When it begins, $k$-BAT calculates and records the prediction performance of all classifiers on the recent batch according to the given \texttt{metric} (this can be any signal about performance such as accuracy, f1-score, etc.). Subsequently, for each classifier, linear regression is performed over the most recent $w$ batches and the slope of the result is recorded (a positive slope indicates that a classifier is generally improving, while a negative slope indicates the reverse).

After recording the performance, delegation begins. The $k$ classifiers with the best performance trend are chosen to act as gurus. Each remaining classifier chooses a single other classifier to whom they delegate. Each delegation is made according to the given \texttt{delegation probability function}, which may use accuracy, weight, or other information to determine the probability that each classifier will delegate to each other classifier.

% Note that $k$-BAT only operates during \textit{training}; during test time, all classifiers in the ensemble are active. The intuition behind this relates to the way we make predictions by summing predicted class probabilities: Typically, classifiers trained on a class have high confidence in their predictions while classifiers not trained on a class have low confidence and randomly distributed predictions, allowing the highly confident trained classifier to dominate.
Note that $k$-BAT only operates during \textit{training}; during test time, all classifiers in the ensemble are active. The idea is that when making predictions, classifiers trained on a specific context are more confident and accurate when predicting classes from that context, while those not trained on that context are less confident and spread their probability mass more evenly between the classes. This means the accurate, confident classifiers mainly determine the final prediction on classes from the context they were trained on.

Below, we describe the individual delegation probability functions that we explore.

\begin{algorithm}[tb]
   \caption{Best Accuracy Trend Delegation Mechanism}
   \label{alg:dynamic_bat}
\begin{algorithmic}

\STATE {\bfseries Input:} $V$, $w$, \texttt{metric}, \texttt{delegation\_probability\_function}

\STATE $P \gets n \times n$ matrix initialized to $0$ 
\FOR{$v_i \in V$}
    % \STATE Do linear regression on accuracy over the last $w$ batches and find the slope, $m_i$, and store in $M$
    \STATE $q_i \gets \text{linear regression slope of } \texttt{metric} \text{ over last } w \text{ batches.}$
\ENDFOR
\FOR{$v_i \in V$}
    \FOR{each $v_{j} \in V$ \textbf{where} $i \neq j$}
        \IF{$q_i > q_j$ and $d(v_i) = v_j$}
            \STATE $d(v_i) := v_i$
        \ELSE
            \STATE $P_{i,j} := $ \texttt{delegation\_probability\_function}$(v_i, v_j)$
            % \STATE $P_{i,j} = \frac{e^{m_j - m_i}}{\sum_{v_k:m_k > m_i}{e^{m_k -m_i}}}$
            % \STATE \ben{Pij needs updating to allow generic, parameterized functions}
        \ENDIF
    \ENDFOR
    \STATE Select a $v_j$ according to probabilities $P_{i,-}$
    \STATE $d(v_i) := v_j$
\ENDFOR
\end{algorithmic}
\end{algorithm}

\subsection{Delegation Probability Functions}

\begin{definition}

The \textbf{Random Better} Delegation Probability Function assigns delegation probabilities such that each classifier with a higher regression slope than the delegator is selected with equal probability.
%selects a delegate for each delegator uniformly at random from voters with strictly higher accuracy \cite{kahng2018liquid}.

\[
p^\text{rand\_better}(v_i, v_j) = \begin{cases} 
      \frac{1}{|N^+(v_i)|} & j \in N^+(v_i) \\
      0 & \text{otherwise}
   \end{cases}
\]

\end{definition}

\begin{definition}

The \textbf{Proportional Better} Delegation Probability Function assigns delegation probabilities such that each classifier with a higher regression slope than the delegator is assigned a probability proportional to their regression slope value. Shown here are delegation probabilities before normalization.

% The \textbf{Proportional Better} Delegation Probability Function has delegators delegate to a representative with higher accuracy, however, the chance of delegating to a voter is directly correlated with the difference between their accuracy and that of the delegator. Shown here are delegation probabilities before normalization.

\[
p^\text{prop\_better}(v_i, v_j) \propto \begin{cases} 
      \frac{q_j - q_i}{\sum_{v_k \in N^+(v_i)} q_k-q_i } & j \in N^+(v_i) \\
      0 & \text{otherwise}
   \end{cases}
\]
\end{definition}

\begin{definition}

The \textbf{Proportional Weighted} Delegation Probability Function assigns delegation probabilities such that each classifier with a higher regression slope than the delegator is assigned a probability proportional to both their regression slope value and the weight of the delegate's guru. A lower weight leads to a higher delegation probability. Shown here are delegation probabilities before normalization.

% The \textbf{Proportional Weighted} Delegation Probability Function returns delegation probabilities based on both the accuracy difference between delegator and delegatee, as well as the weight of the representative ultimately being delegated to. A lower weight leads to a higher delegation probability. Shown here are delegation probabilities before normalization.

\[
p^\text{prop\_weighted}(v_i, v_j) \propto \begin{cases} 
      \frac{1}{w_{d^*(j)}} \frac{q_j - q_i}{\sum_{v_k \in N^+(v_i)} q_k-q_i } & j \in N^+(v_i) \\
      0 & \text{otherwise}
   \end{cases}
\]

\end{definition}

\begin{definition}\label{max-diversity}

The \textbf{Max Diversity} Delegation Probability Function assigns delegation probabilities such that each delegating classifier delegates to the most similar classifier to themselves that has a higher regression slope. Many diversity metrics exist in the ensemble literature~\cite{kuncheva2003measures}; however, in this work, we calculate diversity in a pairwise manner by determining the class prediction probabilities for each classifier (a function easily available for most machine learning models). The ``diversity'' value of any pair of classifiers is the $l2$ norm of the difference between their class probabilities.

\[
p^\text{diverse}(v_i, v_j) = \begin{cases}

      1 & v_j = \argmin_{v_j \in N^+(v_i)} \text{diversity}(v_i, v_j) \\
      0 & \text{otherwise}
   \end{cases}
\]

\end{definition}

\subsection{Student-Expert Delegations}

The Student-Expert extends delegation to test time by running delegation separately on two sets of classifiers: one set which learns (students) and another which is used for inference (experts).
Typically, in continual learning, evaluation is done after training on each context individually or with the assumption that the test stream is drawn i.i.d from all contexts. This can be useful for assessing the degree of forgetting, but in reality, the test stream is likely to be non-stationary, similar to the training stream. We can leverage the non-stationary nature of the test stream with a method that dynamically shifts which members of the ensemble are making predictions. 

The Student-Expert delegation mechanism does just this by adding delegation to the test phase. We use delegation to dynamically shift which members of the ensemble are making predictions. The Student-Expert delegation mechanism can be seen as an extension of the $k$-Best-Accuracy-Trend ($k$-BAT) delegation mechanism described in \autoref{k-BAT}. It maintains $k_s$ ``student'' classifiers and $k_e$ ``expert'' classifiers. $k-BAT$ is run normally for each batch of data to pick $k_s$ students who learn on that batch. Separately, $k-BAT$ is run to pick $k_e$ experts, which will make predictions on the batch. Currently, we use only the Proportional Better delegation probability function to select gurus. Students measure performance using a parameterized window size of $w$ and a given performance metric as in $k$-BAT, while experts measure performance by comparing classifier prediction accuracy on only the most recent batch.

\section{Experiments}

\begin{figure}[ht]
  \centering
  \includegraphics[width=0.95\linewidth]{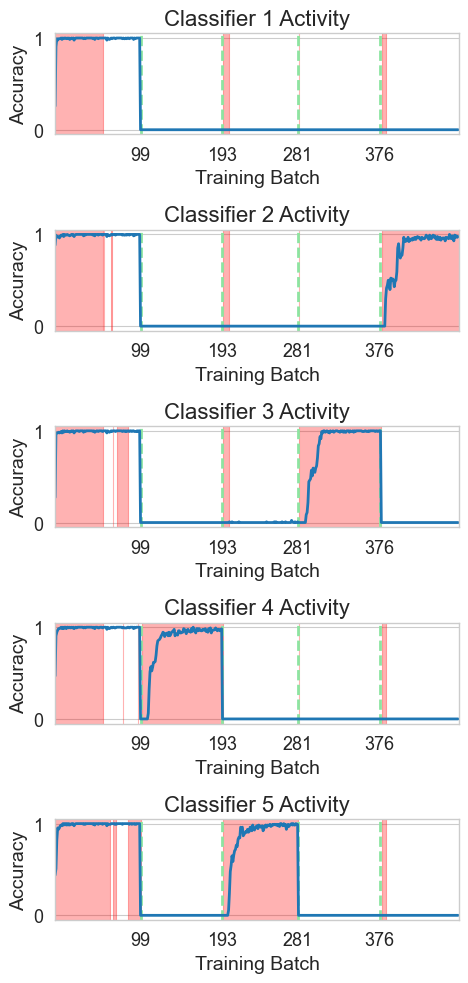}
  \caption{Active learning periods of each classifier on Split MNIST using $k$-BAT with $k = 1$ and probabilistic better delegation. Red periods indicate a classifier is actively learning. With a window size of 50, all classifiers learn on the first 50 batches. Subsequently, delegation begins, and one classifier learns at a time. At context shifts (green lines) various classifiers briefly begin to learn but $k$-BAT effectively picks a single classifier to learn the majority of a context without any knowledge of the context label.}
  \label{fig:fig1}
  \Description{}
\end{figure}

\begin{figure}[ht]
  \centering
  \includegraphics[width=0.99\linewidth]{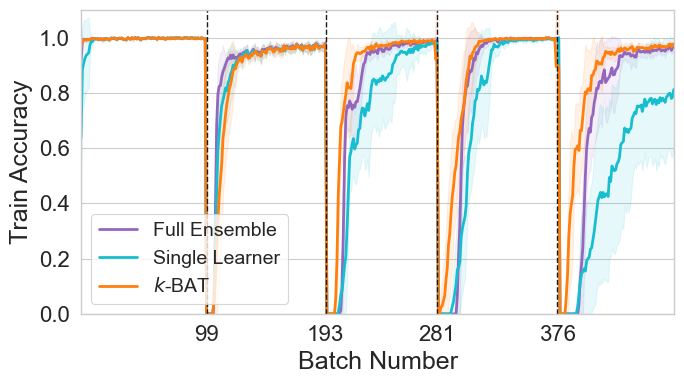}
  \caption{Training accuracy on MNIST of $k$-BAT ($k$ = 1) using probabilistic delegation, a full ensemble, and a single classifier. Results are averaged over 10 trials across 5 disjoint contexts. In early contexts, all models learn at similar rates and achieve similar performance. Later, $k$-BAT continues to learn new contexts quickly while the learning rate of both other models drops.
}
  \label{fig:fig2}
  \Description{}
\end{figure}

We evaluate our algorithms along several axes -- number of gurus, best performance metric, and delegation mechanism. To do this, we run three experiments: First, a simple experiment demonstrates in detail the behaviour of an ensemble using $k$-BAT within a class-incremental setting. Second, a more in-depth experiment explores the behaviour of a variety of parameterizations of our algorithms for class-incremental learning. Finally, a similar experiment explores the same algorithms applied to domain-incremental learning. Our results show strong performance across both settings and surprising differences in optimal parameters for class- and domain-incremental learning.

\subsection{Experimental Setup}

\textbf{Data:}
As is common in continual learning work, we use two MNIST datasets in our experiments: Split MNIST and Rotated MNIST \cite{van2022three}. Split MNIST consists of grey-scale images of digits from 0 to 9, with all images scaled to the same dimensions \cite{zenke2017continual}. The feature values are pixels, and the class is the digit value.
We use Split MNIST as a stream of data partitioned into multiple contexts, each containing a mutually exclusive subset of the classes. 
Specifically, we partition the stream into five disjoint contexts, each containing two adjacent digits (i.e. \{(0, 1), (2, 3), ..., (8, 9)\}. 
We exclusively use Split MNIST in the class-incremental setting where we are interested in learning the global label. On this data, the global label is simply the digit value (0-9).

In order to test our methods in the domain-incremental learning setting, we use a similar but slightly more complex data set: Rotated MNIST. This data again consists of images of digits, as in Split MNIST. However, the digits are now rotated around the center of the image from anywhere between $180^{\circ}$ and $-180^{\circ}$.
We initialize Rotated MNIST into a stream consisting of 5 disjoint contexts. Each context corresponds to a specific rotation amount and contains only images corresponding to that rotation. The context label is the rotation amount, and the within-context label is the digit value (0-9). Each context's rotation is sampled uniformly at random from between $180^{\circ}$ and $-180^{\circ}$.

% \todo{I think this subsubsection can likely be deleted?}
% This section describes the details of the experiments we have run to evaluate our continual learning methods. We ran one basic experiment to illustrate the properties of our $k$-BAT algorithm in a class-incremental learning scenario using the split MNIST dataset. We also ran larger-scale experiments in the class-incremental and domain-incremental settings using the split MNIST and rotated MNIST datasets, respectively. The two larger-scale experiments allowed us to compare our methods to other continual learning baselines.
% For all experiments, results are averaged over 10 trials. Similar to Doan et al. \cite{doan_continual_2023}, we use a single training epoch for all training runs to emulate an online learning setting.

\textbf{Parameter Values:}
We explore a range of parameters in both learning settings. All four delegation probability functions for $k$-BAT are tested for $k \in \{1, 2, 3, 4\}$. We test each delegation probability function and report the best one. Results for other probability delegation functions are shown in the appendix. Additionally, we test three classifier competency metrics for $k$-BAT: accuracy, balanced accuracy, and F1 score. In addition to this, we use between 1 and 4 experts in Student-Expert delegation. Experiments all use ensembles of size $8$ except for the larger experiment on class-incremental learning, which explored an ensemble of size $30$ with up to $11$ gurus, and the smaller illustrative experiment, which used an ensemble size of $5$.

Our models are all simple feed-forward neural networks with two hidden layers which use the Adam optimizer, implemented using Pytorch \cite{paszke2019pytorch}. 
In line with previous work that compares ensemble methods with multiple learners to methods which only use a single learner, we keep the total number of trainable parameters for each continual learning method roughly equal~\cite{wang_coscl_2022}. 
That is, when comparing the performance of an ensemble of 8 classifiers to a single network, each network in the ensemble has roughly one-eighth the number of trainable parameters as the solitary network.

In the class-incremental setting, $k$-BAT and the Student-Expert method use a window size, $w$, of 50 batches. In the domain-incremental setting, these methods use a window size of 400 batches. In all cases, we use a batch size of 128.

\textbf{Benchmarks:} 
We compare our algorithms with three existing approaches to continual learning. A significant component of our algorithms is that they do not explicitly remember previously seen examples (as in Replay-based continual learning) or get told context identities. As such, the methods we compare against also do not use Replay or context identities. Each method we use is found in the Avalanche Continual Learning library \cite{lomonaco2021avalanche}. Conceptually each method is similar, however, their implementation details differ:

\noindent
\textbf{Elastic Weight Consolidation \cite{kirkpatrick2017overcoming}:} EWC slows down updates on network weights that are more important to previously learned examples.

\noindent
\textbf{Learning without Forgetting \cite{li2017learning}:} LwF is similar to EWC but evaluates itself before and after learning new data to try to maintain predictions on previous data.

\noindent
\textbf{Synaptic Intelligence \cite{zenke2017continual}:} Synaptic Intelligence uses biologically motivated techniques to record the importance of each network weight to prediction. Weights important to particular tasks are changed by smaller amounts.s

% In continual learning, it is assumed that the training data will be non-stationary, but usually, the test set is i.i.d over all contexts or performance is evaluated using a test set for each context. While this is useful for assessing the degree of catastrophic forgetting, we may also be able to gain performance by selectively choosing which classifier predictions are being used to make the final classification. In this vein, the Expert Gate strategy \cite{aljundi_expert_2017} dynamically routes test samples to an appropriate member of the ensemble. In line with this, our Student-Expert method is evaluated on a non-stationary version of the test set where each context seen in training is presented again in a stream without the context label available. This allows the Student-Expert method to adjust which classifiers are actively making predictions dynamically. 

\subsection{Results}

\begin{figure}[]
  \centering
  \includegraphics[width=0.99\linewidth]{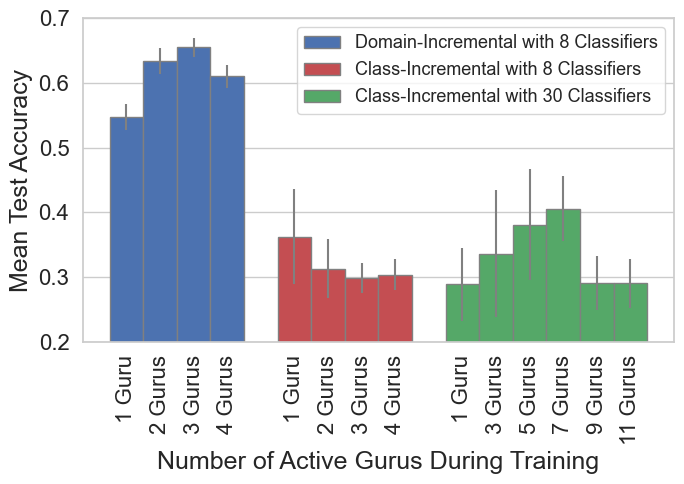}
    \caption{Test accuracy according to value of $k$ for $k$-BAT on domain-incremental learning and two sizes of ensemble doing class-incremental learning. Results are averaged over the probability delegation functions and delegation metrics. While the small-scale experiment on class-incremental learning performs best with only 1 guru, the other settings are optimized with larger numbers of gurus.}
  \label{fig:num_gurus}
  \Description{}
\end{figure}

\begin{table*}[t]
\centering
\scalebox{1}{
\begin{tabular}{@{}lcccccc@{}}
\toprule
Method         & Mean Acc.                 & $C_1$                     & $C_2$                     & $C_3$                     & $C_4$                     & $C_5$                     \\ \hline
Student-Expert & $91.75 \pm 0.39$          & $93.05 \pm 0.85$          & $89.95 \pm 1.07$          & $92.08 \pm 0.42$          & $92.90 \pm 0.15$          & $90.79 \pm 0.33$          \\ \hline
GEM            & \textbf{95.78 $\pm$ 0.22} & $98.39 \pm 0.23$          & \textbf{95.19 $\pm$ 0.65} & \textbf{94.98 $\pm$ 0.76} & \textbf{95.79 $\pm$ 0.35} & $94.55 \pm 0.33$          \\
Replay         & $72.95 \pm 1.76$          & $80.85 \pm 3.00$          & $60.96 \pm 4.77$          & $51.45 \pm 6.30$          & $74.20 \pm 4.73$          & \textbf{97.27 $\pm$ 0.35} \\
MIR            & $52.42 \pm 10.00$         & $60.00 \pm 21.01$         & $57.34 \pm 14.61$         & $64.48 \pm 22.05$         & $0.12 \pm 0.21$           & $80.16 \pm 14.99$         \\ \hline
$k$-BAT        & $43.27 \pm 3.38$          & $73.18 \pm 14.16$         & $2.45 \pm 5.10$           & $93.28 \pm 4.34$          & $35.88 \pm 17.49$         & $11.54 \pm 17.10$         \\
CWR            & $19.98 \pm 0.01$          & \textbf{99.88 $\pm$ 0.05} & $0.00 \pm 0.00$           & $0.00 \pm 0.00$           & $0.00 \pm 0.00$           & $0.00 \pm 0.00$           \\
LwF            & $19.37 \pm 0.14$          & $0.00 \pm 0.00$           & $0.00 \pm 0.00$           & $0.00 \pm 0.00$           & $0.00 \pm 0.00$           & $96.87 \pm 0.70$          \\
MAS            & $25.26 \pm 3.18$          & $18.04 \pm 11.09$         & $1.99 \pm 2.87$           & $0.34 \pm 0.57$           & $9.48 \pm 8.93$           & $96.47 \pm 0.26$          \\
RWalk          & $19.43 \pm 0.06$          & $0.00 \pm 0.00$           & $0.00 \pm 0.00$           & $0.00 \pm 0.00$           & $0.00 \pm 0.00$           & $97.17 \pm 0.28$          \\
Single Learner & $19.44 \pm 0.05$          & $0.00 \pm 0.00$           & $0.00 \pm 0.00$           & $0.00 \pm 0.00$           & $0.00 \pm 0.00$           & $97.19 \pm 0.26$          \\
Full Ensemble  & $19.28 \pm 0.08$          & $0.00 \pm 0.00$           & $0.00 \pm 0.00$           & $0.00 \pm 0.00$           & $0.00 \pm 0.00$           & $96.40 \pm 0.41$          \\ \bottomrule
\end{tabular}
}
\caption{Test accuracy on Split MNIST for all methods averaged over 10 trials in the class-incremental learning setting evaluated after training on all contexts. Results are grouped into 3 sections; the top section shows the performance possible without memory and when leveraging structure within the test set. The middle section compares against pre-existing methods that rely on memory. The bottom section compares methods that do not use memory. Our methods show that ensembles are able to provide significant performance boosts when leveraging structure in test data (Student-Expert with 1 train, 1 test guru), as well as when not memorizing previous examples or being given the context label ($k$-BAT with $k = 1$, Probabilistic Better delegation probability function and balanced accuracy as the performance metric).}
\label{tab:split-mnist-results}
\end{table*}

\begin{table*}[t]
\centering
\scalebox{1}{
\begin{tabular}{@{}lcccccc@{}}
\toprule
Method         & Mean Acc.                 & $C_1$                     & $C_2$                     & $C_3$                     & $C_4$                     & $C_5$                     \\ \midrule
Student Expert & $92.46 \pm 1.39$          & $93.59 \pm 0.62$          & $89.21 \pm 3.27$          & $92.34 \pm 2.41$          & $92.12 \pm 3.94$          & $95.05 \pm 0.68$          \\ \midrule
GEM            & \textbf{95.41 $\pm$ 0.60} & \textbf{94.66 $\pm$ 1.44} & \textbf{94.93 $\pm$ 1.50} & \textbf{94.72 $\pm$ 0.90} & \textbf{95.71 $\pm$ 0.81} & $97.05 \pm 0.21$          \\
Replay         & $75.25 \pm 8.73$          & $64.86 \pm 16.41$         & $71.25 \pm 21.33$         & $64.59 \pm 16.54$         & $78.40 \pm 11.79$         & $97.13 \pm 0.36$          \\
MIR            & $61.97 \pm 11.91$         & $44.83 \pm 24.31$         & $57.07 \pm 32.44$         & $48.97 \pm 23.19$         & $61.83 \pm 22.91$         & $97.15 \pm 0.31$          \\ \midrule
$k$-BAT        & $74.78 \pm 7.24$          & $73.92 \pm 19.46$         & $64.51 \pm 25.03$         & $65.59 \pm 18.22$         & $85.08 \pm 9.95$          & $84.82 \pm 8.49$          \\
CWR            & $55.00 \pm 8.06$          & $65.32 \pm 16.09$         & $43.12 \pm 19.88$         & $43.83 \pm 14.78$         & $62.34 \pm 17.54$         & $60.40 \pm 13.44$         \\
LwF            & $70.81 \pm 9.06$          & $56.92 \pm 21.97$         & $62.23 \pm 30.19$         & $59.89 \pm 20.91$         & $77.81 \pm 14.57$         & \textbf{97.20 $\pm$ 0.27} \\
MAS            & $65.78 \pm 7.57$          & $62.66 \pm 19.10$         & $58.79 \pm 21.73$         & $51.29 \pm 17.00$         & $72.80 \pm 13.26$         & $83.38 \pm 4.10$          \\
RWalk          & $67.77 \pm 9.77$          & $53.48 \pm 21.63$         & $62.51 \pm 28.99$         & $54.77 \pm 21.32$         & $71.48 \pm 16.10$         & $96.61 \pm 0.39$          \\
Single Learner & $62.29 \pm 11.36$         & $45.83 \pm 24.37$         & $57.09 \pm 32.70$         & $48.97 \pm 22.85$         & $62.51 \pm 21.16$         & $97.04 \pm 0.18$          \\
Full Ensemble  & $59.42 \pm 11.86$         & $42.10 \pm 25.15$         & $54.10 \pm 33.59$         & $45.58 \pm 22.74$         & $59.09 \pm 23.05$         & $96.23 \pm 0.44$          \\ \bottomrule
\end{tabular}
}
\caption{Test accuracy for all methods averaged over 10 trials in the domain-incremental learning setting using the Rotated MNIST dataset. All methods are evaluated after training on all contexts. $k$-BAT uses the random better delegation probability function and F1 score as the performance metric and 2 gurus during training. Student-Expert uses the probabilistic better delegation probability function and accuracy as the metric with 2 classifiers active during training and testing. Most methods show strong test performance on the most recently learned data while exhibiting moderate forgetfulness on previously learned data. $k$-BAT exhibits much stronger performance than existing methods that do not rely on replay. The Student-Expert algorithm (top row) highlights the performance possible on an ordered test stream, even without memory.}
\label{tab:rotated-mnist-results}
\end{table*}

In our first experiment, we construct a slightly simplified scenario using the split MNIST data set to illustrate how the $k$-BAT method works. To do so, we plot the individual learning curves of each classifier as well as the periods in which they were active on the training set. Figure \ref{fig:fig1} shows that, roughly speaking, the classifiers divide and conquer, with roughly one classifier learning each context. This is non-trivial since we do not know the index at which the context changes or use the context label to inform which classifier is learning. This is achieved by using the slope of the learning curve over the last $w$ batches. Classifiers that have learned on fewer samples are usually more plastic \cite{lyle_understanding_2023} and, thus, have a steeper learning curve, making them more likely to receive a delegation. Then, when the context switches, the accuracy of a classifier that was learning the previous context will significantly drop, and the slope of the learning curve will eventually become negative. With a negative slope, this classifier will likely delegate to another classifier. Further, since all classifiers that were not learning in the previous context will likely not experience a large drop in accuracy, the slope of their learning curves will be, on average, zero, and thus, they will rescind their delegation to the classifier that was learning in the previous context.

The initial period where all classifiers are active can be explained by the fact that the classifiers do not delegate until their window is full, so by design, all classifiers will be active on the first $w$ batches. Since not all classifiers are learning in each context, the classifiers specialize in a reduced number of contexts and thus, the catastrophic forgetting is greatly reduced.

In addition to better performance on the test set, \autoref{fig:fig2} shows that in later contexts, the $k$-BAT Delegation ensemble learns more quickly, which is likely because the classifiers learning in the BAT delegation ensemble have been trained for fewer iterations than the classifiers in the full ensemble, and thus have higher plasticity~\cite{lyle_understanding_2023}.

\subsubsection{Class-Incremental Learning}
In our second experiment, we perform a more thorough investigation of the $k$-BAT and Student-Expert methods using various parameter combinations in our delegative framework. In particular, we vary the number of active learning gurus during training for both the $k$-BAT and the Student-Expert method, and we vary the number of active predicting gurus for the Student-Expert method. We also vary the delegation probability function and the classifier competency metric for $k$-BAT.

% \autoref{tab:split-mnist-results} shows the average accuracy of each method on the full test set as well as the average accuracy in each context on the Split MNIST dataset. Both $k$-BAT and the Student-Expert method significantly outperform all baselines. The Student-Expert method is the only one to leverage order in the test set so its performance far surpasses all methods that predict statically on the test set. Order in the test set is, of course, not always present, so we present these results separately in \autoref{tab:split-mnist-results}.
% However, even $k$-BAT, which does not receive any batch labels during testing is much more accurate during testing than the Full Ensemble, Single Learner, LwF, EWC and the Synaptic Intelligence methods.

\autoref{tab:split-mnist-results} shows the average accuracy of each method on the full test set as well as the average accuracy in each context on the Split MNIST dataset. Both $k$-BAT and the Student-Expert method significantly outperform naive methods (Full Ensemble and Single Learner). The Student-Expert method is the only one to leverage order in the test set and thus significantly outperforms all similar methods that do not rely on memory. Order in the test set is rarely available, so we separate these results from others.
However, even $k$-BAT, which does not receive any batch labels during testing, is more accurate than other methods that do not rely on memory.

% Looking at the performance on each of the five contexts reveals that $k$-BAT and Student-Expert are the only methods which remember any information from the first four contexts, indicating strong ability to avoid catastrophic forgetting. In regard to methods which predict on the whole test set statically, our $k$-BAT method with $1$ active learning guru performs particularly well, achieving greater than zero accuracy in each context.

The performance on each of the five contexts reveals that $k$-BAT and Student-Expert successfully remember information from the first four contexts, indicating a strong ability to avoid catastrophic forgetting. In regard to methods which predict on the whole test set statically, our $k$-BAT method with $1$ active learning guru performs particularly well, achieving greater than zero accuracy in each context.

% The Student-Expert model shows the dramatic benefit to learnability that comes with structure in the test set. Overall, we see that Student-Expert with $1$ guru during both training and testing achieved a $90.5\%$ test accuracy while $k$-BAT with $1$ guru achieved only a $38.6\%$ accuracy. However, even $k$-BAT is much higher than all other methods which achieve only $~19.5\%$ accuracy -- corresponding to excellent performance on one-fifth of the data (the final context).

The Student-Expert model shows the benefit that comes with structure in an online test stream. Overall, we see that Student-Expert with $1$ guru during both training and testing achieved a $91.75\%$ test accuracy while $k$-BAT with $1$ guru achieved only a $43.27\%$ accuracy. However, even $k$-BAT is much higher than all other methods that similarly don't rely on memory.

\autoref{fig:num_gurus} shows that, in the larger class-incremental setting with $30$ classifiers, having as many as $7$ gurus can maximize performance. However, in smaller ensembles, $1$ guru appears optimal.
When varying the delegation probability function and classifier competency metric in $k$-BAT, we found no significant differences between any of them. A comparison of test accuracy across these parameters is shown in Appendix \ref{sec:appendix}.

% We also look at how various parameters, such as the delegation probability function and the classifier competency metric, affect the performance of our $k$-BAT method. \autoref{fig:num_gurus} demonstrates how the number of active learning gurus, $k$, impacts performance. In the class-incremental setting with $n=8$ classifiers in the ensemble, having only one classifier learning at a time results in the best performance. In contrast to this, with $n=30$, having seven classifiers actively learning at a time results in the best performance. We also found that in the class-incremental setting, the choice between using the slope of the accuracy, balanced accuracy, or F1, metric in $k$-BAT, has nearly no impact on performance (\autoref{fig:metric_comparison}). As for the delegation probability function used in $k$-BAT, max-diversity, as described in \textit{definition} \ref{max-diversity}, performs slightly better than the other options.

\subsubsection{Domain-Incremental Learning}
% In our third experiment, we investigate the performance of $k$-BAT and Student-Expert in the domain-incremental setting. The results of this experiment can be seen in \autoref{tab:rotated-mnist-results}. 

In our third experiment, we investigate the performance of $k$-BAT and Student-Expert in the domain-incremental setting. The results of this experiment can be seen in \autoref{tab:rotated-mnist-results}. 
The Student-Expert method significantly outperforms all other methods that do not use memory, achieving an average test accuracy of $92.46\%$ with 2 active gurus during training and testing. This is primarily due to the fact that Student-Expert operates on an ordered, online test stream, as discussed previously.
Of the remaining methods that don't use memory, $k$-BAT with $k = 2$ achieves the highest average accuracy of $74.78\%$.

% The Student-Expert method significantly outperforms all other methods, achieving an average test accuracy of $88.4\%$ with 2 student gurus during training and 3 expert gurus during testing. This is primarily due to the fact that Student-Expert operates on an ordered test stream, as discussed previously.
% Of the remaining methods, $k$-BAT with $k = 2$ achieves the highest average accuracy of $67.1\%$. However, $k$-BAT with 1, 3, or 4 gurus also outperforms all other methods tested. In this setting LwF outperforms the full ensemble, single learner, EWC, and Synaptic Intelligence, offering a strong $57.3\%$ average test accuracy.

Impressively, both Student-Expert and $k$-BAT perform similarly well in the first context as in some later contexts. Overall, both of our methods avoid significant forgetting.

% Impressively, both Student-Expert and $k$-BAT perform similarly well in the first context as in some later contexts. Overall both of our methods avoid significant forgetting, especially compared to the benchmark methods which are themselves an improvement over a full ensemble and single learner.

When averaging over all parameter configurations for $k$-BAT, using $k = 3$ resulted in the best performance. Similar to our findings in the class-incremental setting, each performance metric leads to similar performance. Of the delegation probability functions, max-diversity slightly outperforms the other options, but the difference is insignificant. These results can be seen in Appendix \ref{sec:appendix}.

\section{Discussion}

In this paper, we have developed new ensemble training algorithms based on delegative techniques for two types of continual learning: class-incremental and domain-incremental.
Our first algorithm, $k$-BAT, shows a strong ability to select a unique ensemble member to learn each new context, or domain shift, without any knowledge of the context itself or the context label. This effectively teaches different members of the ensemble on different distributions and drastically reduces the catastrophic forgetting common to continual learning settings.
Subsequently, our Student-Expert delegation mechanism extends $k$-BAT to allow delegation during testing to improve performance in the presence of ordered test data. This manages to avoid catastrophic forgetting and showcases the power of delegation for dynamically selecting classifiers to both learn and make predictions.

Curiously, in our class-incremental learning experiments with only 8 classifiers in the ensemble, we found that 1 guru led to the highest accuracy. This is not the case with domain-incremental learning, where multiple gurus are superior. Intuitively, when there are more classifiers than contexts, having multiple classifiers learn on a context should have better performance than when just one classifier is learning. Our result suggests that either (1) individual classifiers are quite strong in the class-incremental setting and that more gurus may be useful when the learning task is more difficult, or (2) there is room for improvement in the delegation algorithm. In either case, further investigation will be useful.

Generally, our experiments have varied several important parameters to demonstrate both the optimal parameter values and the robustness of our algorithms. However, there is a great deal of work that could extend this research in compelling directions.
Some specific paths include:

\begin{itemize}
	\item Optimal delegations: We have no guarantee of the quality of our delegative process. Both theoretical and further experimental work will be important to identify an upper bound on the performance improvement granted by our delegation mechanisms.
	\item New performance metrics for delegation: Our delegation mechanisms accept a generic signal of quality as a means of deciding delegation patterns. Measuring the change in this value over time acts as an approximation of each classifier's learning rate. While we experimented with a number of common metrics such as accuracy and f1-score, there may be metrics that more directly reveal which classifiers are optimal to learn on any given data.
	\item Further comparison with existing methods to avoid catastrophic forgetting in continual learning. We compared our results with three methods well suited for domain-incremental learning, however these methods prove to work poorly on the class-incremental setting.
\end{itemize}

% Interesting observations to expand upon:

% \begin{itemize}
%     \item Often only one guru is better than many. Probably suggests lots of room for future work but also due to the nature of this setting.
%     \item Delegation is useful dynamic ensemble selection but the transitive aspect of LD is probably not necessary.
%     \item Why is 1 guru good for class incremental and more than one good for domain incremental?
% \end{itemize}

% \subsection{Future Work}

% \begin{itemize}
%     \item Measures of performance that evaluate how ``saturated'' with knowledge a voter is? Or, more generally, that are specifically geared towards determining who is best to learn some context.
%      \item We can possibly incorporate some single learner continual learning methods into the members of the ensembles
% \end{itemize}

%%%%%%%%%%%%%%%%%%%%%%%%%%%%%%%%%%%%%%%%%%%%%%%%%%%%%%%%%%%%%%%%%%%%%%%%

%%% The next two lines define, first, the bibliography style to be 
%%% applied, and, second, the bibliography file to be used.

\bibliographystyle{ACM-Reference-Format} 
\bibliography{sample}


\begin{thebibliography}{25}

%%% the Web address.

\ifx \showCODEN    \undefined \def \showCODEN     #1{\unskip}     \fi
\ifx \showDOI      \undefined \def \showDOI       #1{#1}\fi
\ifx \showISBNx    \undefined \def \showISBNx     #1{\unskip}     \fi
\ifx \showISBNxiii \undefined \def \showISBNxiii  #1{\unskip}     \fi
\ifx \showISSN     \undefined \def \showISSN      #1{\unskip}     \fi
\ifx \showLCCN     \undefined \def \showLCCN      #1{\unskip}     \fi
\ifx \shownote     \undefined \def \shownote      #1{#1}          \fi
\ifx \showarticletitle \undefined \def \showarticletitle #1{#1}   \fi
\ifx \showURL      \undefined \def \showURL       {\relax}        \fi
% invisible to TeX
\providecommand\bibfield[2]{#2}
\providecommand\bibinfo[2]{#2}
\providecommand\natexlab[1]{#1}
\providecommand\showeprint[2][]{arXiv:#2}

\bibitem[\protect\citeauthoryear{Aljundi, Chakravarty, and Tuytelaars}{Aljundi et~al\mbox{.}}{2017}]%
        {aljundi_expert_2017}
\bibfield{author}{\bibinfo{person}{Rahaf Aljundi}, \bibinfo{person}{Punarjay Chakravarty}, {and} \bibinfo{person}{Tinne Tuytelaars}.} \bibinfo{year}{2017}\natexlab{}.
\newblock \showarticletitle{Expert {Gate}: {Lifelong} {Learning} {With} a {Network} of {Experts}}. \bibinfo{pages}{3366--3375}.
\newblock
\urldef\tempurl%
\url{https://openaccess.thecvf.com/content_cvpr_2017/html/Aljundi_Expert_Gate_Lifelong_CVPR_2017_paper.html}
\showURL{%
\tempurl}


\bibitem[\protect\citeauthoryear{Armstrong and Larson}{Armstrong and Larson}{2021}]%
        {armstrong2021limited}
\bibfield{author}{\bibinfo{person}{Ben Armstrong} {and} \bibinfo{person}{Kate Larson}.} \bibinfo{year}{2021}\natexlab{}.
\newblock \showarticletitle{On the Limited Applicability of Liquid Democracy}.
\newblock \bibinfo{journal}{\emph{3rd Games, Agents, and Incentives Workshop at the 20th International Conference on Autonomous Agents and Multiagent Systems (AAMAS)}} (\bibinfo{year}{2021}).
\newblock


\bibitem[\protect\citeauthoryear{Armstrong and Larson}{Armstrong and Larson}{2024}]%
        {armstrong2024liquid}
\bibfield{author}{\bibinfo{person}{Ben Armstrong} {and} \bibinfo{person}{Kate Larson}.} \bibinfo{year}{2024}\natexlab{}.
\newblock \bibinfo{title}{Liquid Democracy for Low-Cost Ensemble Pruning}.
\newblock
\newblock
\showeprint{2401.17443}


\bibitem[\protect\citeauthoryear{Chaudhry, Ranzato, Rohrbach, and Elhoseiny}{Chaudhry et~al\mbox{.}}{2019}]%
        {chaudhry2018efficient}
\bibfield{author}{\bibinfo{person}{Arslan Chaudhry}, \bibinfo{person}{Marc’Aurelio Ranzato}, \bibinfo{person}{Marcus Rohrbach}, {and} \bibinfo{person}{Mohamed Elhoseiny}.} \bibinfo{year}{2019}\natexlab{}.
\newblock \showarticletitle{Efficient Lifelong Learning with A-{GEM}}. In \bibinfo{booktitle}{\emph{International Conference on Learning Representations}}.
\newblock
\urldef\tempurl%
\url{https://openreview.net/forum?id=Hkf2_sC5FX}
\showURL{%
\tempurl}


\bibitem[\protect\citeauthoryear{Cornelio, Donini, Loreggia, Pini, and Rossi}{Cornelio et~al\mbox{.}}{2021}]%
        {cornelio2021voting}
\bibfield{author}{\bibinfo{person}{Cristina Cornelio}, \bibinfo{person}{Michele Donini}, \bibinfo{person}{Andrea Loreggia}, \bibinfo{person}{Maria~Silvia Pini}, {and} \bibinfo{person}{Francesca Rossi}.} \bibinfo{year}{2021}\natexlab{}.
\newblock \showarticletitle{Voting with Random Classifiers (VORACE): Theoretical and Experimental Analysis}.
\newblock \bibinfo{journal}{\emph{Autonomous Agents and Multi-Agent Systems}} \bibinfo{volume}{35}, \bibinfo{number}{2} (\bibinfo{year}{2021}), \bibinfo{pages}{22}.
\newblock


\bibitem[\protect\citeauthoryear{Doan, Mirzadeh, and Farajtabar}{Doan et~al\mbox{.}}{2023}]%
        {doan_continual_2023}
\bibfield{author}{\bibinfo{person}{Thang Doan}, \bibinfo{person}{Seyed~Iman Mirzadeh}, {and} \bibinfo{person}{Mehrdad Farajtabar}.} \bibinfo{year}{2023}\natexlab{}.
\newblock \bibinfo{title}{Continual {Learning} {Beyond} a {Single} {Model}}.
\newblock
\newblock
\urldef\tempurl%
\url{https://doi.org/10.48550/arXiv.2202.09826}
\showDOI{\tempurl}
\newblock
\shownote{arXiv:2202.09826 [cs].}


\bibitem[\protect\citeauthoryear{French}{French}{1999}]%
        {french1999catastrophic}
\bibfield{author}{\bibinfo{person}{Robert~M French}.} \bibinfo{year}{1999}\natexlab{}.
\newblock \showarticletitle{Catastrophic Forgetting in Connectionist Networks}.
\newblock \bibinfo{journal}{\emph{Trends in cognitive sciences}} \bibinfo{volume}{3}, \bibinfo{number}{4} (\bibinfo{year}{1999}), \bibinfo{pages}{128--135}.
\newblock


\bibitem[\protect\citeauthoryear{Freund and Schapire}{Freund and Schapire}{1995}]%
        {freund1995decision}
\bibfield{author}{\bibinfo{person}{Yoav Freund} {and} \bibinfo{person}{Robert~E Schapire}.} \bibinfo{year}{1995}\natexlab{}.
\newblock \showarticletitle{A Decision-Theoretic Generalization of On-line Learning and an Application to Boosting}. In \bibinfo{booktitle}{\emph{European conference on computational learning theory}}. Springer, \bibinfo{pages}{23--37}.
\newblock


\bibitem[\protect\citeauthoryear{Kirkpatrick, Pascanu, Rabinowitz, Veness, Desjardins, Rusu, Milan, Quan, Ramalho, Grabska-Barwinska, et~al\mbox{.}}{Kirkpatrick et~al\mbox{.}}{2017}]%
        {kirkpatrick2017overcoming}
\bibfield{author}{\bibinfo{person}{James Kirkpatrick}, \bibinfo{person}{Razvan Pascanu}, \bibinfo{person}{Neil Rabinowitz}, \bibinfo{person}{Joel Veness}, \bibinfo{person}{Guillaume Desjardins}, \bibinfo{person}{Andrei~A Rusu}, \bibinfo{person}{Kieran Milan}, \bibinfo{person}{John Quan}, \bibinfo{person}{Tiago Ramalho}, \bibinfo{person}{Agnieszka Grabska-Barwinska}, {et~al\mbox{.}}} \bibinfo{year}{2017}\natexlab{}.
\newblock \showarticletitle{Overcoming Catastrophic Forgetting in Neural Networks}.
\newblock \bibinfo{journal}{\emph{Proceedings of the national academy of sciences}} \bibinfo{volume}{114}, \bibinfo{number}{13} (\bibinfo{year}{2017}), \bibinfo{pages}{3521--3526}.
\newblock


\bibitem[\protect\citeauthoryear{Kuncheva and Whitaker}{Kuncheva and Whitaker}{2003}]%
        {kuncheva2003measures}
\bibfield{author}{\bibinfo{person}{Ludmila~I Kuncheva} {and} \bibinfo{person}{Christopher~J Whitaker}.} \bibinfo{year}{2003}\natexlab{}.
\newblock \showarticletitle{Measures of Diversity in Classifier Ensembles and their Relationship with the Ensemble Accuracy}.
\newblock \bibinfo{journal}{\emph{Machine learning}}  \bibinfo{volume}{51} (\bibinfo{year}{2003}), \bibinfo{pages}{181--207}.
\newblock


\bibitem[\protect\citeauthoryear{Leon, Floria, and B{\u{a}}dic{\u{a}}}{Leon et~al\mbox{.}}{2017}]%
        {leon2017evaluating}
\bibfield{author}{\bibinfo{person}{Florin Leon}, \bibinfo{person}{Sabina-Adriana Floria}, {and} \bibinfo{person}{Costin B{\u{a}}dic{\u{a}}}.} \bibinfo{year}{2017}\natexlab{}.
\newblock \showarticletitle{Evaluating the Effect of Voting Methods on Ensemble-Based Classification}. In \bibinfo{booktitle}{\emph{2017 IEEE International Conference on INnovations in Intelligent Systems and Applications}}. IEEE, \bibinfo{pages}{1--6}.
\newblock


\bibitem[\protect\citeauthoryear{Li and Hoiem}{Li and Hoiem}{2017}]%
        {li2017learning}
\bibfield{author}{\bibinfo{person}{Zhizhong Li} {and} \bibinfo{person}{Derek Hoiem}.} \bibinfo{year}{2017}\natexlab{}.
\newblock \showarticletitle{Learning Without Forgetting}.
\newblock \bibinfo{journal}{\emph{IEEE transactions on pattern analysis and machine intelligence}} \bibinfo{volume}{40}, \bibinfo{number}{12} (\bibinfo{year}{2017}), \bibinfo{pages}{2935--2947}.
\newblock


\bibitem[\protect\citeauthoryear{Liu, Parisot, Slabaugh, Jia, Leonardis, and Tuytelaars}{Liu et~al\mbox{.}}{2020}]%
        {liu_more_2020}
\bibfield{author}{\bibinfo{person}{Yu Liu}, \bibinfo{person}{Sarah Parisot}, \bibinfo{person}{Gregory Slabaugh}, \bibinfo{person}{Xu Jia}, \bibinfo{person}{Ales Leonardis}, {and} \bibinfo{person}{Tinne Tuytelaars}.} \bibinfo{year}{2020}\natexlab{}.
\newblock \showarticletitle{More {Classifiers}, {Less} {Forgetting}: {A} {Generic} {Multi}-classifier {Paradigm} for {Incremental} {Learning}}. In \bibinfo{booktitle}{\emph{Computer {Vision} – {ECCV} 2020}} \emph{(\bibinfo{series}{Lecture {Notes} in {Computer} {Science}})}, \bibfield{editor}{\bibinfo{person}{Andrea Vedaldi}, \bibinfo{person}{Horst Bischof}, \bibinfo{person}{Thomas Brox}, {and} \bibinfo{person}{Jan-Michael Frahm}} (Eds.). \bibinfo{publisher}{Springer International Publishing}, \bibinfo{address}{Cham}, \bibinfo{pages}{699--716}.
\newblock
\showISBNx{978-3-030-58574-7}
\urldef\tempurl%
\url{https://doi.org/10.1007/978-3-030-58574-7_42}
\showDOI{\tempurl}


\bibitem[\protect\citeauthoryear{Lomonaco, Pellegrini, Cossu, Carta, Graffieti, Hayes, De~Lange, Masana, Pomponi, Van~de Ven, et~al\mbox{.}}{Lomonaco et~al\mbox{.}}{2021}]%
        {lomonaco2021avalanche}
\bibfield{author}{\bibinfo{person}{Vincenzo Lomonaco}, \bibinfo{person}{Lorenzo Pellegrini}, \bibinfo{person}{Andrea Cossu}, \bibinfo{person}{Antonio Carta}, \bibinfo{person}{Gabriele Graffieti}, \bibinfo{person}{Tyler~L Hayes}, \bibinfo{person}{Matthias De~Lange}, \bibinfo{person}{Marc Masana}, \bibinfo{person}{Jary Pomponi}, \bibinfo{person}{Gido~M Van~de Ven}, {et~al\mbox{.}}} \bibinfo{year}{2021}\natexlab{}.
\newblock \showarticletitle{Avalanche: An End-to-End Library for Continual Learning}. In \bibinfo{booktitle}{\emph{Proceedings of the IEEE/CVF Conference on Computer Vision and Pattern Recognition}}. \bibinfo{pages}{3600--3610}.
\newblock


\bibitem[\protect\citeauthoryear{Lyle, Zheng, Nikishin, Pires, Pascanu, and Dabney}{Lyle et~al\mbox{.}}{2023}]%
        {lyle_understanding_2023}
\bibfield{author}{\bibinfo{person}{Clare Lyle}, \bibinfo{person}{Zeyu Zheng}, \bibinfo{person}{Evgenii Nikishin}, \bibinfo{person}{Bernardo~Avila Pires}, \bibinfo{person}{Razvan Pascanu}, {and} \bibinfo{person}{Will Dabney}.} \bibinfo{year}{2023}\natexlab{}.
\newblock \showarticletitle{Understanding {Plasticity} in {Neural} {Networks}}. In \bibinfo{booktitle}{\emph{Proceedings of the 40th {International} {Conference} on {Machine} {Learning}}}. \bibinfo{publisher}{PMLR}, \bibinfo{pages}{23190--23211}.
\newblock
\urldef\tempurl%
\url{https://proceedings.mlr.press/v202/lyle23b.html}
\showURL{%
\tempurl}
\newblock
\shownote{ISSN: 2640-3498.}


\bibitem[\protect\citeauthoryear{Pan, Swaroop, Immer, Eschenhagen, Turner, and Khan}{Pan et~al\mbox{.}}{2020}]%
        {pan_continual_2020}
\bibfield{author}{\bibinfo{person}{Pingbo Pan}, \bibinfo{person}{Siddharth Swaroop}, \bibinfo{person}{Alexander Immer}, \bibinfo{person}{Runa Eschenhagen}, \bibinfo{person}{Richard Turner}, {and} \bibinfo{person}{Mohammad Emtiyaz~E Khan}.} \bibinfo{year}{2020}\natexlab{}.
\newblock \showarticletitle{Continual {Deep} {Learning} by {Functional} {Regularisation} of {Memorable} {Past}}. In \bibinfo{booktitle}{\emph{Advances in {Neural} {Information} {Processing} {Systems}}}, Vol.~\bibinfo{volume}{33}. \bibinfo{publisher}{Curran Associates, Inc.}, \bibinfo{pages}{4453--4464}.
\newblock
\urldef\tempurl%
\url{https://proceedings.neurips.cc/paper/2020/hash/2f3bbb9730639e9ea48f309d9a79ff01-Abstract.html}
\showURL{%
\tempurl}


\bibitem[\protect\citeauthoryear{Paszke, Gross, Massa, Lerer, Bradbury, Chanan, Killeen, Lin, Gimelshein, Antiga, et~al\mbox{.}}{Paszke et~al\mbox{.}}{2019}]%
        {paszke2019pytorch}
\bibfield{author}{\bibinfo{person}{Adam Paszke}, \bibinfo{person}{Sam Gross}, \bibinfo{person}{Francisco Massa}, \bibinfo{person}{Adam Lerer}, \bibinfo{person}{James Bradbury}, \bibinfo{person}{Gregory Chanan}, \bibinfo{person}{Trevor Killeen}, \bibinfo{person}{Zeming Lin}, \bibinfo{person}{Natalia Gimelshein}, \bibinfo{person}{Luca Antiga}, {et~al\mbox{.}}} \bibinfo{year}{2019}\natexlab{}.
\newblock \showarticletitle{Pytorch: An Imperative sStyle, High-performance Deep Learning Library}.
\newblock \bibinfo{journal}{\emph{Advances in neural information processing systems}}  \bibinfo{volume}{32} (\bibinfo{year}{2019}).
\newblock


\bibitem[\protect\citeauthoryear{Pennock, Maynard-Reid~II, Giles, and Horvitz}{Pennock et~al\mbox{.}}{2000}]%
        {pennock2000normative}
\bibfield{author}{\bibinfo{person}{David~M Pennock}, \bibinfo{person}{Pedrito Maynard-Reid~II}, \bibinfo{person}{C~Lee Giles}, {and} \bibinfo{person}{Eric Horvitz}.} \bibinfo{year}{2000}\natexlab{}.
\newblock \showarticletitle{A Normative Examination of Ensemble Learning Algorithms.}. In \bibinfo{booktitle}{\emph{ICML}}. \bibinfo{pages}{735--742}.
\newblock


\bibitem[\protect\citeauthoryear{Rebuffi, Kolesnikov, Sperl, and Lampert}{Rebuffi et~al\mbox{.}}{2017}]%
        {rebuffi_icarl_2017}
\bibfield{author}{\bibinfo{person}{Sylvestre-Alvise Rebuffi}, \bibinfo{person}{Alexander Kolesnikov}, \bibinfo{person}{Georg Sperl}, {and} \bibinfo{person}{Christoph~H. Lampert}.} \bibinfo{year}{2017}\natexlab{}.
\newblock \showarticletitle{{iCaRL}: {Incremental} {Classifier} and {Representation} {Learning}}. In \bibinfo{booktitle}{\emph{2017 {IEEE} {Conference} on {Computer} {Vision} and {Pattern} {Recognition} ({CVPR})}}. \bibinfo{publisher}{IEEE}, \bibinfo{address}{Honolulu, HI}, \bibinfo{pages}{5533--5542}.
\newblock
\showISBNx{978-1-5386-0457-1}
\urldef\tempurl%
\url{https://doi.org/10.1109/CVPR.2017.587}
\showDOI{\tempurl}


\bibitem[\protect\citeauthoryear{Rypeść, Cygert, Khan, Trzciński, Zieliński, and Twardowski}{Rypeść et~al\mbox{.}}{2024}]%
        {rypesc_divide_2024}
\bibfield{author}{\bibinfo{person}{Grzegorz Rypeść}, \bibinfo{person}{Sebastian Cygert}, \bibinfo{person}{Valeriya Khan}, \bibinfo{person}{Tomasz Trzciński}, \bibinfo{person}{Bartosz Zieliński}, {and} \bibinfo{person}{Bartłomiej Twardowski}.} \bibinfo{year}{2024}\natexlab{}.
\newblock \bibinfo{title}{Divide and not Forget: {Ensemble} of Selectively Trained Experts in {Continual} {Learning}}.
\newblock
\newblock
\urldef\tempurl%
\url{http://arxiv.org/abs/2401.10191}
\showURL{%
\tempurl}
\newblock
\shownote{arXiv:2401.10191 [cs].}


\bibitem[\protect\citeauthoryear{van~de Ven, Tuytelaars, and Tolias}{van~de Ven et~al\mbox{.}}{2022}]%
        {van2022three}
\bibfield{author}{\bibinfo{person}{Gido~M van~de Ven}, \bibinfo{person}{Tinne Tuytelaars}, {and} \bibinfo{person}{Andreas~S Tolias}.} \bibinfo{year}{2022}\natexlab{}.
\newblock \showarticletitle{Three Types of Incremental Learning}.
\newblock \bibinfo{journal}{\emph{Nature Machine Intelligence}} \bibinfo{volume}{4}, \bibinfo{number}{12} (\bibinfo{year}{2022}), \bibinfo{pages}{1185--1197}.
\newblock


\bibitem[\protect\citeauthoryear{Wang, Zhang, Li, Zhu, and Zhong}{Wang et~al\mbox{.}}{2022}]%
        {wang_coscl_2022}
\bibfield{author}{\bibinfo{person}{Liyuan Wang}, \bibinfo{person}{Xingxing Zhang}, \bibinfo{person}{Qian Li}, \bibinfo{person}{Jun Zhu}, {and} \bibinfo{person}{Yi Zhong}.} \bibinfo{year}{2022}\natexlab{}.
\newblock \showarticletitle{{CoSCL}: {Cooperation} of {Small} {Continual} {Learners} is {Stronger} {Than} a {Big} {One}}. In \bibinfo{booktitle}{\emph{Computer {Vision} – {ECCV} 2022}} \emph{(\bibinfo{series}{Lecture {Notes} in {Computer} {Science}})}, \bibfield{editor}{\bibinfo{person}{Shai Avidan}, \bibinfo{person}{Gabriel Brostow}, \bibinfo{person}{Moustapha Cissé}, \bibinfo{person}{Giovanni~Maria Farinella}, {and} \bibinfo{person}{Tal Hassner}} (Eds.). \bibinfo{publisher}{Springer Nature Switzerland}, \bibinfo{address}{Cham}, \bibinfo{pages}{254--271}.
\newblock
\showISBNx{978-3-031-19809-0}
\urldef\tempurl%
\url{https://doi.org/10.1007/978-3-031-19809-0_15}
\showDOI{\tempurl}


\bibitem[\protect\citeauthoryear{Wang, Zhang, Su, and Zhu}{Wang et~al\mbox{.}}{2023}]%
        {wang2023comprehensive}
\bibfield{author}{\bibinfo{person}{Liyuan Wang}, \bibinfo{person}{Xingxing Zhang}, \bibinfo{person}{Hang Su}, {and} \bibinfo{person}{Jun Zhu}.} \bibinfo{year}{2023}\natexlab{}.
\newblock \showarticletitle{A Comprehensive Survey of Continual Learning: Theory, Method and Application}.
\newblock \bibinfo{journal}{\emph{arXiv preprint arXiv:2302.00487}} (\bibinfo{year}{2023}).
\newblock


\bibitem[\protect\citeauthoryear{Wen, Tran, and Ba}{Wen et~al\mbox{.}}{2020}]%
        {wen_batchensemble_2020}
\bibfield{author}{\bibinfo{person}{Yeming Wen}, \bibinfo{person}{Dustin Tran}, {and} \bibinfo{person}{Jimmy Ba}.} \bibinfo{year}{2020}\natexlab{}.
\newblock \bibinfo{title}{{BatchEnsemble}: {An} {Alternative} {Approach} to {Efficient} {Ensemble} and {Lifelong} {Learning}}.
\newblock
\newblock
\urldef\tempurl%
\url{https://doi.org/10.48550/arXiv.2002.06715}
\showDOI{\tempurl}
\newblock
\shownote{arXiv:2002.06715 [cs, stat].}


\bibitem[\protect\citeauthoryear{Zenke, Poole, and Ganguli}{Zenke et~al\mbox{.}}{2017}]%
        {zenke2017continual}
\bibfield{author}{\bibinfo{person}{Friedemann Zenke}, \bibinfo{person}{Ben Poole}, {and} \bibinfo{person}{Surya Ganguli}.} \bibinfo{year}{2017}\natexlab{}.
\newblock \showarticletitle{Continual Learning Through Synaptic Intelligence}. In \bibinfo{booktitle}{\emph{International conference on machine learning}}. PMLR, \bibinfo{pages}{3987--3995}.
\newblock


\end{thebibliography}

\newpage
\clearpage

\appendix

\section{Additional Experiment Results}
\label{sec:appendix}
\begin{figure}[h]
    \centering
    \includegraphics[width=0.99\linewidth]{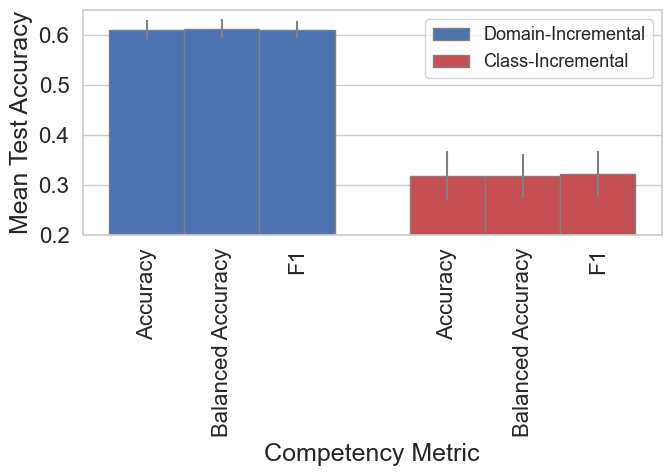}
    \caption{Comparison of performance metrics for both class- and domain-incremental learning. In both cases, there is no significant difference between any metrics we explored.}
    \label{fig:metric_comparison}
\end{figure}

\begin{figure}[h]
\vspace{8mm}
    \centering
    \includegraphics[width=0.99\linewidth]{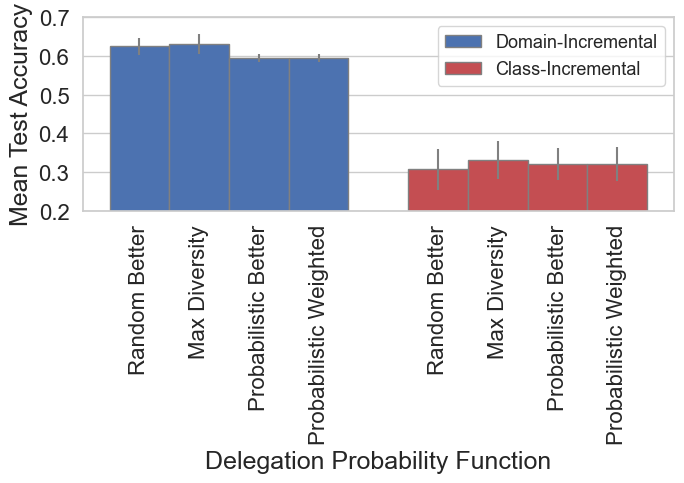}
    \caption{Comparison of delegation probability functions for both class- and domain-incremental learning. While Max Diversity may be slightly better, there is no significant difference between the functions in the experiments we have run.}
    \label{fig:prob_function_comparison}
\end{figure}

\begin{figure*}[ht]
    \centering
    \includegraphics[width=\textwidth]{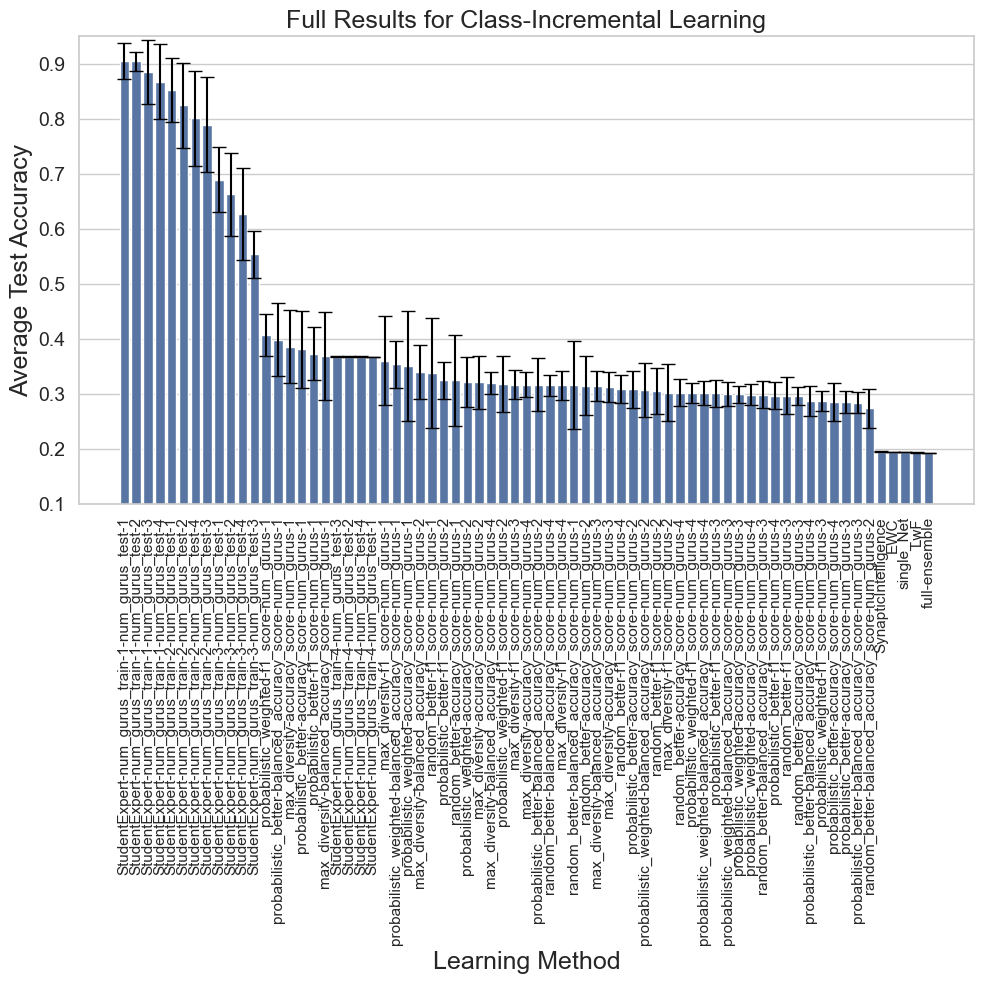}
    \caption{Average test accuracy of all compared methods on class-incremental learning. Student-Expert methods utilize the ordered test set and inevitably outperform other methods. Unlike in domain-incremental learning, methods with only 1 guru typically perform better than others.}
    \label{fig:full_CIL_results}
\end{figure*}

\begin{figure*}[ht]
    \centering
    \includegraphics[width=\textwidth]{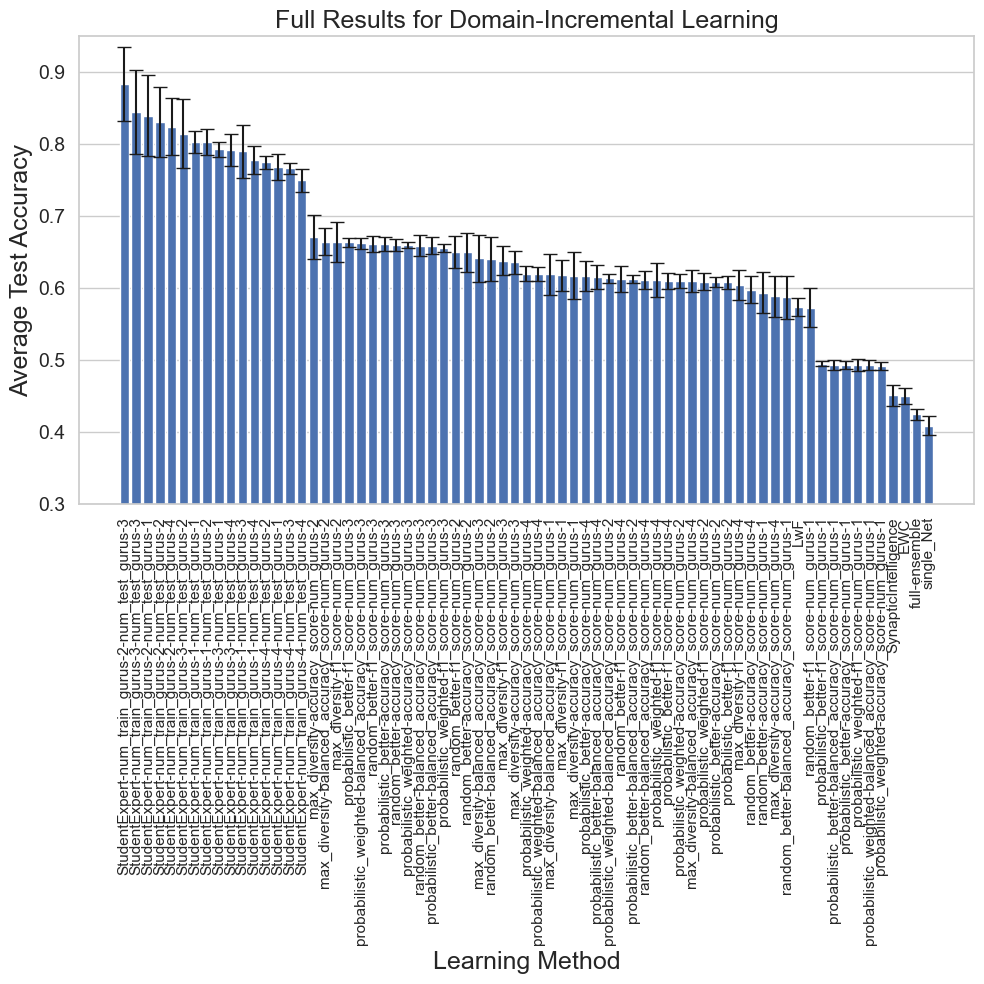}
    \caption{Average test accuracy of all compared methods on domain-incremental learning. Student-Expert methods utilize the ordered test set and inevitably outperform other methods. Unlike in class-incremental learning, methods with more than one guru typically perform better than others.}
    \label{fig:full_DIL_results}
\end{figure*}

%%%%%%%%%%%%%%%%%%%%%%%%%%%%%%%%%%%%%%%%%%%%%%%%%%%%%%%%%%%%%%%%%%%%%%%%

\end{document}